\documentclass{article}

\usepackage{PRIMEarxiv}

\usepackage[utf8]{inputenc}
\usepackage[T1]{fontenc}
\usepackage{hyperref}
\usepackage{url}
\usepackage{booktabs}
\usepackage{amsfonts}
\usepackage{amsmath}
\usepackage{nicefrac}
\usepackage{microtype}
\usepackage{fancyhdr}
\usepackage{graphicx}
\usepackage{booktabs, multirow}
\usepackage{soul}
\usepackage[table]{xcolor}
\usepackage{changepage,threeparttable}
\usepackage{caption}
\graphicspath{{./}}

\makeatletter
\newcommand{\naturalfnmarks}{\renewcommand{\@makefnmark}{\hbox{$^{\@thefnmark}$}}}
\makeatother

\title{A Transformer-Based Contrastive Learning Approach for Few-Shot Sign Language Recognition}

\author{
    \naturalfnmarks
    Silvan Ferreira\thanks{These authors contributed equally to this work.}\thanks{Present affiliation: Universidade Federal do Rio Grande do Norte (UFRN), Natal, Brazil.}, Esdras Costa\footnotemark[1], Marcio Dahia \\
    CESAR \\
    Recife, Brazil \\
    \texttt{\{sfsj, esc, mlmd\}@cesar.org} \\
    \And
    Jampierre Rocha \\
    Lenovo \\
    S\~ao Paulo, Brazil \\
    \texttt{jrocha2@lenovo.com} \\
}

\begin{document}
\maketitle

\begin{abstract}
Sign language recognition from monocular video or 2D pose sequences is challenging, both because 3D information must be inferred from 2D observations and because the signal is inherently spatiotemporal. Moreover, the large and continually growing vocabulary of signs in production settings makes conventional closed-set classification impractical: adding a class requires new labeled data and retraining. We propose a contrastive Transformer-based model that learns rich representations of body key-point sequences, enabling direct comparison between embedding vectors. These representations support one-shot and few-shot tasks such as classification of signs never seen during training. On the LSA64 dataset, using only 48 classes for representation learning, the model reaches 88.4\,$\pm$\,1.8\% accuracy on 16 held-out classes with as few as eight reference examples per class, and its accuracy improves consistently with the number of training classes and support examples. We release the full training and evaluation code, including distributed training and dataset preparation scripts, to enable exact reproduction.\footnote{Code: \url{https://github.com/esdrascosta/slr-fewshot}}
\end{abstract}

\keywords{Sign Language \and Transformers \and Contrastive Learning \and Few-shot Learning}

\section{Introduction}
\label{section:intro}
According to the World Health Organization (WHO), more than 700 million people are expected to have disabling hearing loss by 2050, and over five percent of the global population --- more than 466 million people --- is currently deaf \cite{world2021world}. Deaf people primarily communicate using a family of spatiotemporal languages called sign languages \cite{world2021world}. Each country usually has its own sign language with particular gestures and meanings; Libras (Brazilian Sign Language), for example, is an official language of Brazil \cite{de2012linguistic}. Countries that do not adopt a sign language as an official language often address accessibility through public and educational policies, such as the Individuals with Disabilities Education Act (IDEA) in the United States \cite{schwartz2022falling}.

Sign language remains understudied, and its gestural nature makes communication with non-signers difficult. This language barrier arises because deaf people frequently do not master the surrounding written language, while only few hearing people can sign \cite{barros2016traduccao}. Learners who were born deaf typically face great difficulty acquiring reading and writing skills \cite{da2017qlibras}, since sign languages differ markedly from written languages in grammar, morphology, syntax, and semantics, often expressing the same idea through completely different structures \cite{rocha2020towards}.

Many studies seek to overcome this communication barrier, most of them framing the problem as computer vision with machine learning. Progress is limited, however, by existing sign language datasets, which cover a small number of words with few examples per word \cite{li2020word}.

To address these problems, we propose a pipeline consisting of a pose model that extracts body and hand key-points for each frame, followed by a Transformer encoder that models the temporal structure of the key-point sequence of the entire input video. Unlike recurrent units such as LSTMs or GRUs, which consume the input step by step, the Transformer processes the whole sequence in parallel, and its self-attention mechanism handles long sequences effectively.

The model is trained with a contrastive objective, the triplet loss, after which it maps key-point sequences to semantically meaningful vectors in an embedding space. Classification can then be performed by comparing embeddings with simple similarity rules such as nearest neighbors or cosine similarity. Consequently, a new input can be classified against only a few examples of each known sign, and new classes can be added without any retraining.

The main contribution of this paper is a method for few-shot classification of isolated signs that uses a Transformer encoder trained with a contrastive objective, together with a systematic evaluation of three few-shot classification rules (Prototypical Networks, $k$-nearest neighbors, and cosine similarity) on classes never seen during training. We additionally release a fully reproducible, distributed-training-ready implementation.

The remainder of the paper is organized as follows: Section~\ref{section:related} reviews related work; Section~\ref{section:model} describes the proposed model; Section~\ref{section:training} details the training procedure and implementation; Section~\ref{section:results} presents the experiments and results; and Section~\ref{section:concl} concludes and discusses future work.

\section{Related Work}
\label{section:related}
Sign Language Recognition (SLR) can be divided into two main areas: \textbf{single sign recognition}, which targets signs representing one concept or gloss, and \textbf{continuous sign recognition}, which deals with continuous streams of signs forming sentences. This paper addresses single sign recognition.

Different approaches have been used to address sign recognition \cite{cooper2011sign}. One of the first works was published by \cite{murakami1991gesture}, who used recurrent neural networks (RNNs) to recognize single signs from finger-alphabet symbol sequences, but could not perform temporal sign segmentation since training was done at the sign level. \cite{kim1996dynamic} and \cite{lee1997real} trained Fuzzy Min-Max Neural Networks (FMMNNs) with 25 single signs and 131 words, respectively. Other neural approaches include \cite{waldron1995isolated}, which combined layers trained at different levels of sign abstraction, and \cite{yang2002extraction}, which used 2D motion trajectories to train a Time-Delay Neural Network (TDNN) for American Sign Language (ASL).

In the 1990s, Hidden Markov Models dominated the field \cite{starner1997real,grobel1997isolated}, given the similarity to speech recognition and their convenient treatment of the temporal dimension. In addition, \cite{kim2001continuous} used Deterministic Finite Automata (DFA) to recognize Korean Sign Language (KSL), while \cite{kadous1996machine} used $k$-nearest neighbors and decision trees to classify single signs.

Since then, deep neural networks have been widely applied to SLR. \cite{pigou2014sign} used Convolutional Neural Networks (CNNs) to classify 20 Italian gestures from Microsoft Kinect full-body images with depth, achieving 91.7\% accuracy. \cite{jhuang2007biologically} proposed a 3D CNN whose third dimension spans video frames for human action recognition, and \cite{huang2015sign} used 3D CNNs to extract spatial and temporal features from multiple visual data sources.

Beyond per-frame visual features, temporal information must be captured across the video. To this end, \cite{zuo2015convolutional} proposed the Convolutional Recurrent Neural Network (C-RNN), combining a CNN with an RNN to model contextual dependencies between images; \cite{bantupalli2018american} applied this architecture to classify signs from an ASL dataset.

Transformer networks \cite{vaswani2017attention} have achieved impressive results in Natural Language Processing (NLP) and computer vision. In NLP, models such as BERT \cite{devlin2018bert} trained on large text corpora reached state-of-the-art results in translation, text classification, and question answering. \cite{girdhar2019video} used Transformers to recognize and localize human actions in video from sequences of RGB frames. Transformers are also present in SLR: \cite{camgoz2020sign} connected a Transformer to a Connectionist Temporal Classification (CTC) objective for continuous SLR, and \cite{jiang2021skeletor} showed robustness to corrupted skeleton key-point detections.

The literature shows that single sign recognition is usually implemented as a supervised, closed-set classification task in which the model is trained with a fixed number of classes and many labeled examples per class. This approach is expensive in terms of data acquisition and does not scale: a new class can only be incorporated by collecting labeled examples and retraining the network. The problem is particularly critical in SLR, where vocabularies commonly contain tens of thousands of distinct signs.

Considering the scalability problem, \cite{shovkopliassupport} investigated few-shot learning techniques --- Matching Networks, Model-Agnostic Meta-Learning, and Prototypical Networks --- on very small datasets of electromyograms of sign performance, obtaining promising results and suggesting the possibility of transfer between sign languages. \cite{bilge2022towards} went further, addressing zero-shot sign language recognition (ZSSLR) by pairing videos of single signs with textual descriptions that act as an intermediate semantic representation for knowledge transfer to unseen signs.

In this paper, we achieve few-shot single sign recognition by encoding key-point sequences with a Transformer trained using the triplet loss, an objective proposed by \cite{schroff2015facenet} that achieved state-of-the-art results in face recognition. The method is detailed in Sections~\ref{section:model} and \ref{section:training}.

\section{Proposed Model}
\label{section:model}
\begin{figure}
  \centering
  \includegraphics[width=16.0cm]{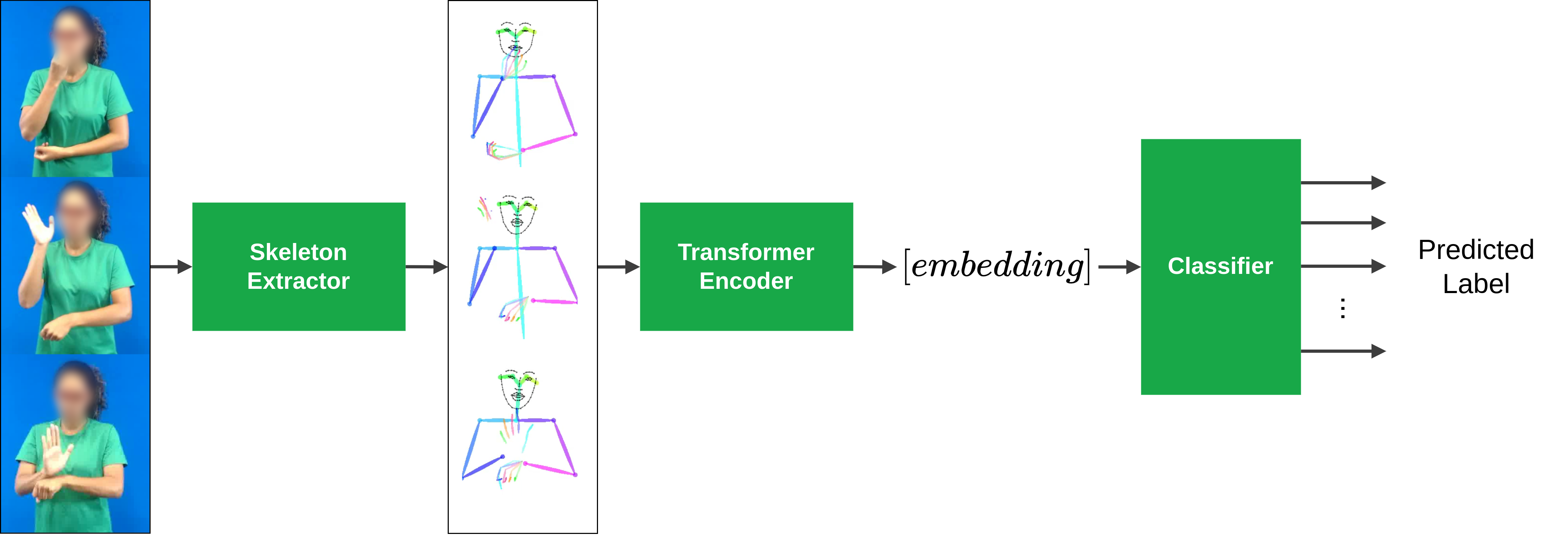}
  \caption{Proposed pipeline overview. RGB images are the input of a pose-and-hands (skeleton) model; the Transformer encoder predicts the embedding representation of the sequence; and the classifier predicts the input class from the embedding.}
  \label{fig:big_picture}
\end{figure}

The proposed pipeline goes from an input video, represented as a sequence of RGB frames, to a predicted output class. First, key-points are extracted from the RGB images by a pose-and-hands model, forming a skeleton of the person performing the sign. The skeleton sequence is passed to the Transformer encoder, which produces an embedding of the whole sequence; finally, a few-shot classification rule predicts the class. An overview is shown in Figure~\ref{fig:big_picture}.

\subsection{Skeleton Extraction}
\label{subsec:skeleton}
The sequence of RGB frames could be fed directly to a model as raw pixels, e.g., using CNNs \cite{o2015introduction} or image Transformers \cite{dosovitskiy2020image}. However, the complexity of raw pixel input demands larger models capable of building good internal representations \cite{bengio2013representation}, and correspondingly more data to prevent overfitting or shortcut learning, where the model exploits spurious characteristics of the training data instead of the true underlying distribution \cite{goodfellow2016deep}. With a limited training set and computational budget, training directly on pixels tends to perform poorly and converge slowly.

For sign language recognition, only the configuration of the body, hands, and face is required; appearance factors contained in the pixels, such as color, shadow, and depth, should not affect classification. It is therefore good practice to use a pre-trained key-point predictor \cite{cao2019openpose,lugaresi2019mediapipe}. Key-points are coordinates representing specific regions of the subject, such as joints, fingers, or facial landmarks.

\begin{figure}
  \centering
  \includegraphics[width=16.0cm]{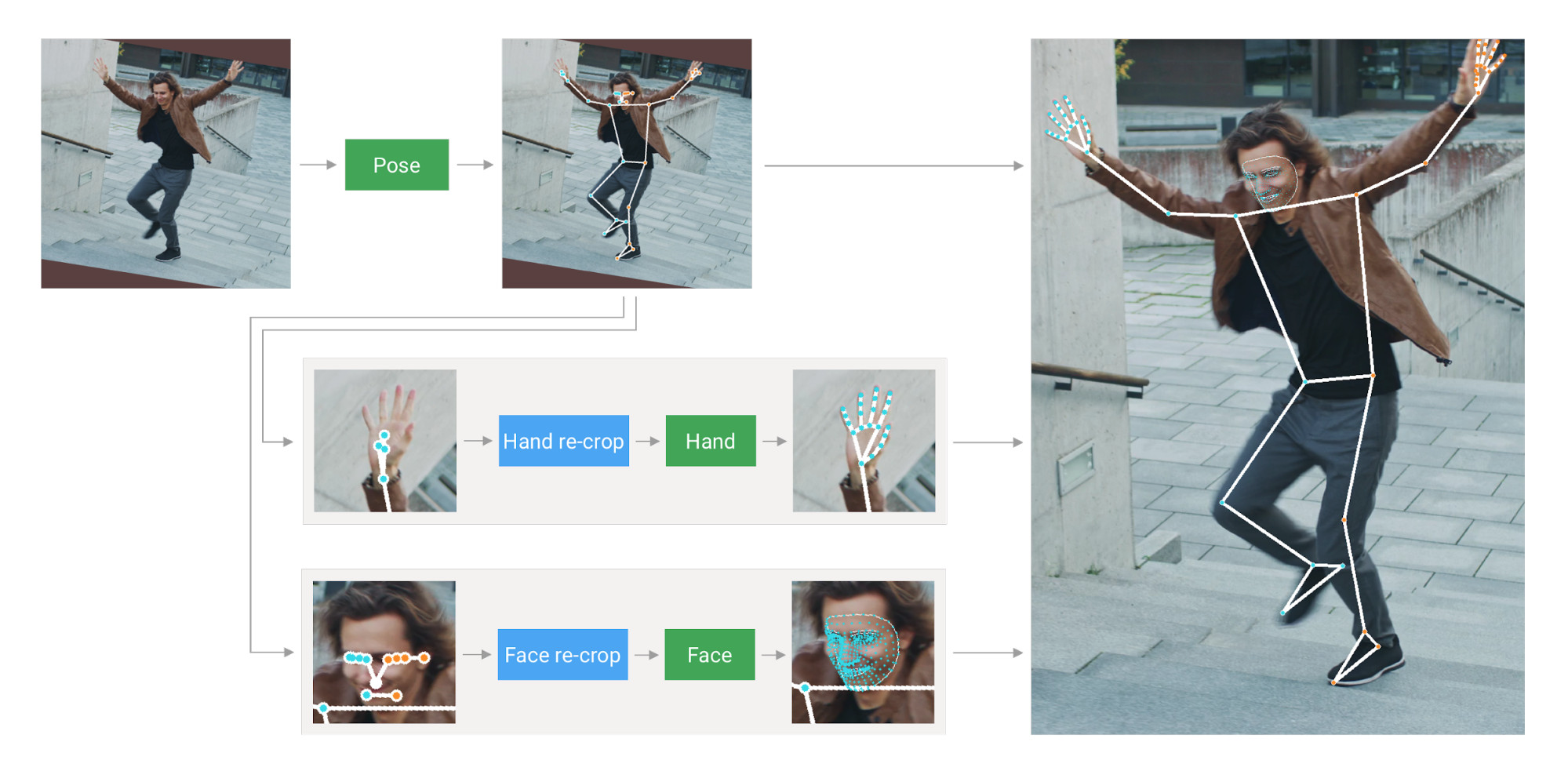}
  \caption{MediaPipe Holistic overview (source: \cite{google2020holistic}).}
  \label{fig:mediapipe}
\end{figure}

In this work, MediaPipe Holistic \cite{lugaresi2019mediapipe} is used to extract skeletons from RGB images. MediaPipe Holistic is a pipeline of separate models for pose, face, and hands (Figure~\ref{fig:mediapipe}), each with its own specialization and input format: the pose model receives a fixed low-resolution image ($256\times256$), which would be too coarse for the hands, so a multi-stage pipeline re-crops each component at an appropriate resolution. The face model is not used in our experiments.

The skeleton is built from the pose (33 key-points) and both hands (21 key-points each), yielding $M=75$ key-points per frame. We retain the three coordinates $(x, y, z)$ predicted by MediaPipe for each key-point, so each frame is described by $3M = 225$ features.\footnote{The $z$ coordinate estimated from monocular input is noisier than $x$ and $y$; our released code allows discarding it, reducing the per-frame dimension to $2M=150$. The experiments reported here keep the $z$ coordinate (\texttt{use\_z=true}).} For a video $V$ with $N$ frames, the extracted sequence is $X=(X_1,\dots,X_N)$, where $X_i=(x^1_i,y^1_i,z^1_i,\dots,x^M_i,y^M_i,z^M_i)\in\mathbb{R}^{3M}$ collects the coordinates of all key-points of the $i$-th frame; hence $X\in\mathbb{R}^{N\times 3M}$.

Additionally, to remove bias related to dynamic aspects of signing, such as speed and camera frame rate, a temporal interpolation standardizes the length of every skeleton sequence to $N$ frames (per-coordinate linear interpolation over a uniform time grid). This preprocessing step is applied in all experiments reported in Section~\ref{section:results}.

\subsection{Transformer}
\begin{figure}
    \centering
    \includegraphics[width=14.0cm]{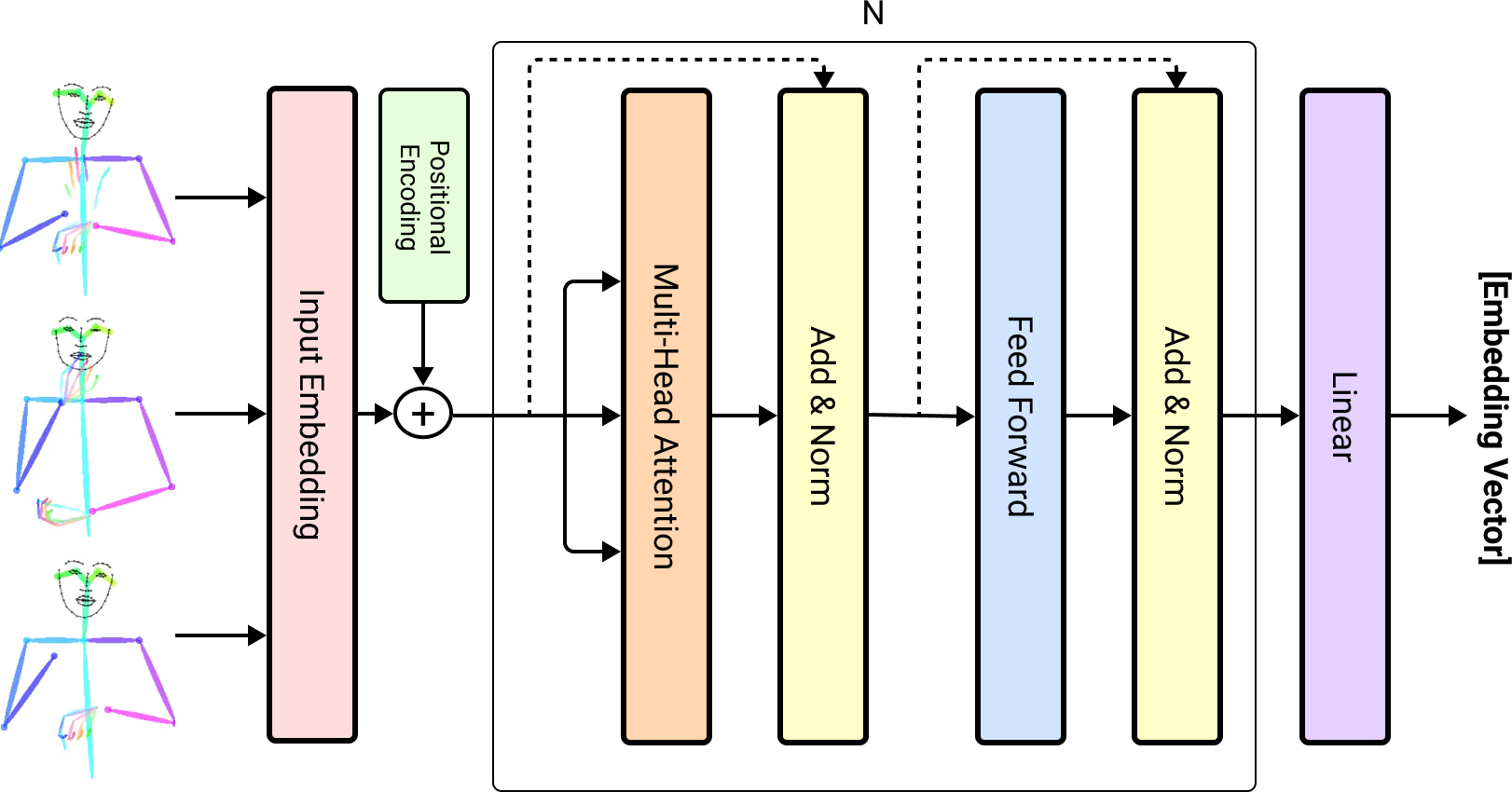}
    \caption{The model receives a sequence of skeleton key-points, creates a vector representation, adds positional information, and the Transformer encoder maps the sequence to an embedding vector through the final hidden state of the \texttt{[CLS]} token followed by a linear layer.}
    \label{fig:transformer}
\end{figure}

The coordinate sequence $X$ serves as input to a multi-layer Transformer encoder based on \cite{vaswani2017attention}. As in BERT \cite{devlin2018bert}, a special classification token, \texttt{[CLS]}, is prepended to every sequence; its final hidden state is used as the aggregate sequence representation and is passed through a final linear layer to produce the embedding used for comparisons (Figure~\ref{fig:transformer}).

\subsubsection{Pose Embedding}
Before entering the encoder, each frame vector $X_t$ is embedded into the $d_{model}$-dimensional space of the Transformer by a single-layer feed-forward network with ReLU activation:
\begin{equation}
    \label{eq:emb}
    X_t^{emb}=\mathrm{ReLU}(\textbf{W}_{e}X_{t}+\textbf{b}_{e})
\end{equation}
where $\textbf{W}_e$ and $\textbf{b}_e$ are learnable parameters that extract features from $X_t\in\mathbb{R}^{3M}$ and produce a representation $X_t^{emb}\in\mathbb{R}^{d_{model}}$ in which frames with similar content lie close together.

Because the Transformer is non-recurrent and consumes the whole sequence in parallel, information about element order must be injected explicitly. A positional encoding vector, computed from periodic functions at frequencies that depend on the position, is added to each input embedding:
\begin{align}
    PE_{(pos, 2i)}&=\sin\!\left(pos/10000^{2i/d_{model}}\right) \label{eq:pos1} \\
    PE_{(pos, 2i+1)}&=\cos\!\left(pos/10000^{2i/d_{model}}\right) \label{eq:pos2}
\end{align}
where $pos$ is the position of the frame in the sequence and $i$ indexes the embedding dimension.

\subsubsection{Transformer Encoder}
The encoder is a stack of $n=2$ layers, each containing two sub-layers: a multi-head self-attention mechanism and a fully connected feed-forward network. Each sub-layer is wrapped in a residual connection followed by layer normalization:
\begin{equation}
    \label{eq:enc}
    \mathrm{Output}=\mathrm{LayerNorm}(x+\mathrm{Sublayer}(x))
\end{equation}
where $\mathrm{Sublayer}$ is the function implemented by the sub-layer. To facilitate the residual connections, all sub-layers produce outputs of dimension $d_{model}=128$.

\subsubsection{Attention}
The self-attention mechanism maps a query and key-value pairs to an output computed as a weighted sum of the values, where each weight is given by a compatibility function between the query and the corresponding key. Each embedding $X_t^{emb}$ is multiplied by learnable matrices $W_Q$, $W_K$, and $W_V$ to produce the packed matrices $Q$, $K$, and $V$:
\begin{equation}
    \label{eq:attn}
    \mathrm{Attention}(Q,K,V)=\mathrm{softmax}\!\left(\frac{QK^T}{\sqrt{d_k}}\right)V
\end{equation}

Instead of a single attention function, several attention heads are computed in parallel, allowing the model to attend to different representation sub-spaces. The outputs are concatenated and projected to form the Multi-Head Attention (MHA) result:
\begin{equation}
    \label{eq:mha}
    \mathrm{MultiHead}(Q,K,V)=\mathrm{Concat}(\mathrm{head}_1,...,\mathrm{head}_h)W_O
\end{equation}
where $\mathrm{head}_i$ is the self-attention output of the $i$-th head and $W_O$ is a learnable matrix mapping the concatenated dimension $h\,d_v$ back to $d_{model}$. We use $h=4$ heads.

\subsubsection{Feed-Forward Networks}
The second sub-layer is a position-wise feed-forward network applying two linear transformations with a ReLU in between:
\begin{equation}
    \label{eq:ff}
    \mathrm{FFN}(x)=\max(0,\, xW_1+b_1)W_2+b_2
\end{equation}
where $W_1$, $W_2$, $b_1$, and $b_2$ are learnable parameters that differ from layer to layer.

\subsection{One-Shot and Few-Shot Classification}
\label{subsec:fewshot}
The embedding of an input video lies in a $d$-dimensional space. The mapping should be semantic: under a given distance metric, vectors of the same class must be close, and vectors of different classes far apart.

Classification then reduces to comparing the embedding of a new input against the embeddings of a few labeled reference examples (the \emph{support set}) of each candidate class; the closest references determine the prediction. Crucially, the support classes may be entirely disjoint from the classes used to train the encoder. We evaluate three classification rules: $k$-nearest neighbors, cosine similarity, and Prototypical Networks.

\subsubsection{$k$-Nearest Neighbors}
\begin{figure}
\centering
\includegraphics[width=6.5 cm]{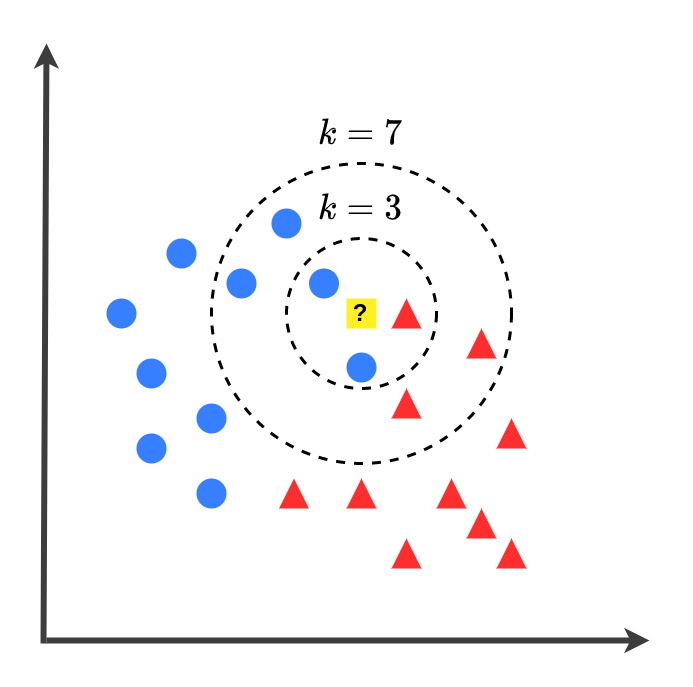}
\caption{Visualization of $k$-nearest neighbors with two classes (blue circles and red triangles) for $k=3$ and $k=7$.}
\label{fig:knn}
\end{figure}

The $k$-nearest neighbors algorithm ($k$NN) \cite{kramer2013k} is a non-parametric method for classification and regression. In classification, it uses the labeled support data and a distance function, usually Euclidean, to compute the distance between each support sample and a given new input, predicting the label from the nearest points (Figure~\ref{fig:knn}).

Consider a support set of $S$ samples, each a pair $(\textbf{x}_j, y_j)$ with $\textbf{x}_j=(x_j^1,\dots,x_j^P)$ a point with $P$ features and $y_j$ its label. The Euclidean distance between two points is
\begin{equation}
    \label{eq:knn_euclidian_dist}
    d(\textbf{x}_j, \textbf{x}_k)=\sqrt{(x_j^1-x_k^1)^2+(x_j^2-x_k^2)^2+\dots+(x_j^P-x_k^P)^2}.
\end{equation}
For a query $\textbf{q}$ and $k=1$, the predicted label is that of the nearest support sample,
\begin{equation}
    \label{eq:knn}
    \hat{y}(\textbf{q}) = y_{j^*}, \qquad j^*=\operatorname*{arg\,min}_{j\in\{1,\dots,S\}} d(\textbf{x}_j, \textbf{q}),
\end{equation}
and for $k>1$ the predicted class is the majority class among the $k$ nearest support samples.

\subsubsection{Cosine Similarity}
\begin{figure}
\centering
\includegraphics[width=14.5 cm]{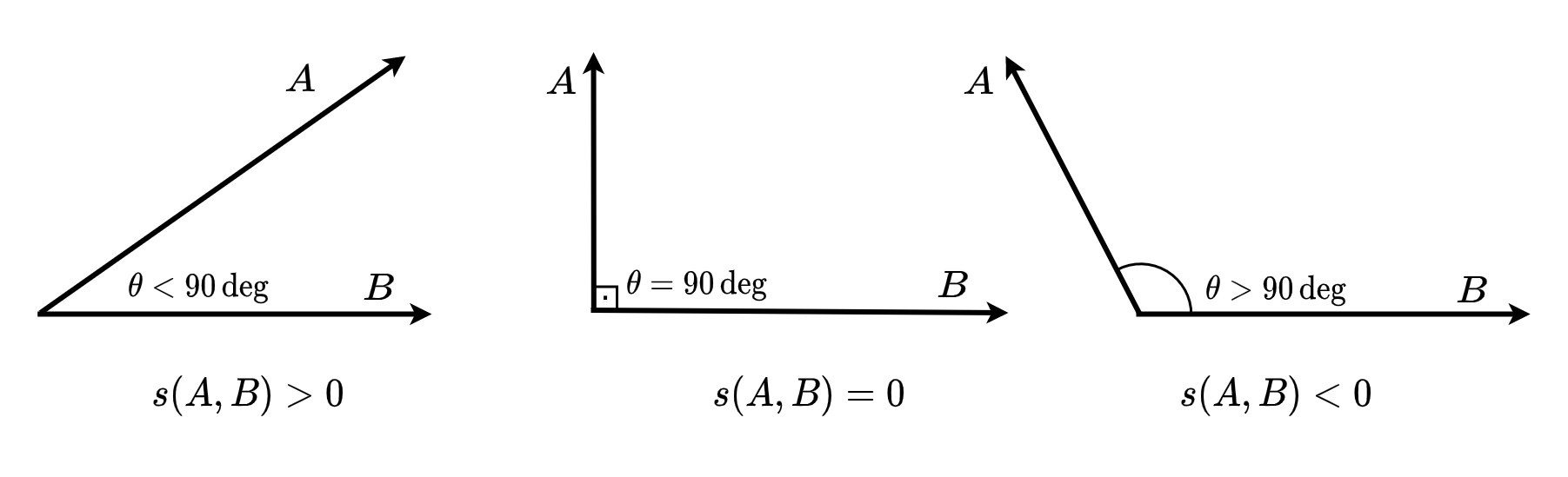}
\caption{Visualization of cosine similarity for $s(\textbf{A},\textbf{B})>0$, $s(\textbf{A},\textbf{B})=0$, and $s(\textbf{A},\textbf{B})<0$, respectively.}
\label{fig:cos}
\end{figure}

Cosine similarity \cite{dehak2010cosine} measures how similar two vectors are in an inner-product space. In Euclidean space, the inner product of $\textbf{A}$ and $\textbf{B}$ is
\begin{equation}
    \label{eq:cos_inner}
    \textbf{A}\cdot\textbf{B}=\sum_i^{P}{a_ib_i},
\end{equation}
which can be interpreted geometrically as the degree of alignment between the vectors (Figure~\ref{fig:cos}): maximal when they point in the same direction and zero when they are perpendicular. Normalizing by the product of the magnitudes yields a value in $[-1, 1]$:
\begin{equation}
    \label{eq:cos_similarity}
    s(\textbf{A},\textbf{B})=\cos(\theta)=\frac{\textbf{A}\cdot\textbf{B}}{\Vert\textbf{A}\Vert\,\Vert\textbf{B}\Vert}.
\end{equation}

As a classifier, the query embedding $\textbf{q}$ is compared with all support samples, and the prediction is the class of the most similar one:
\begin{equation}
    \label{eq:cos}
    \hat{y}(\textbf{q}) = y_{j^*}, \qquad j^*=\operatorname*{arg\,max}_{j\in\{1,\dots,S\}} s(\textbf{x}_j, \textbf{q}).
\end{equation}

\subsubsection{Prototypical Networks}
Prototypical Networks \cite{snell2017prototypical} learn a metric space in which classification is performed by computing distances to a prototype representation of each class. An embedding function $f_{\theta}: \mathbb{R}^D \to \mathbb{R}^d$ with learnable parameters $\theta$ encodes each input; the prototype of class $k$ is the mean of the embedded support points of that class:
\begin{equation}
    \label{eq:proto}
    \textbf{c}_k = \frac{1}{|S_k|} \sum_{(x_i, y_i) \in S_k} f_{\theta}(x_i)
\end{equation}
where $S_k$ denotes the support set of class $k$.

\begin{figure}
    \centering
    \includegraphics[scale=0.35]{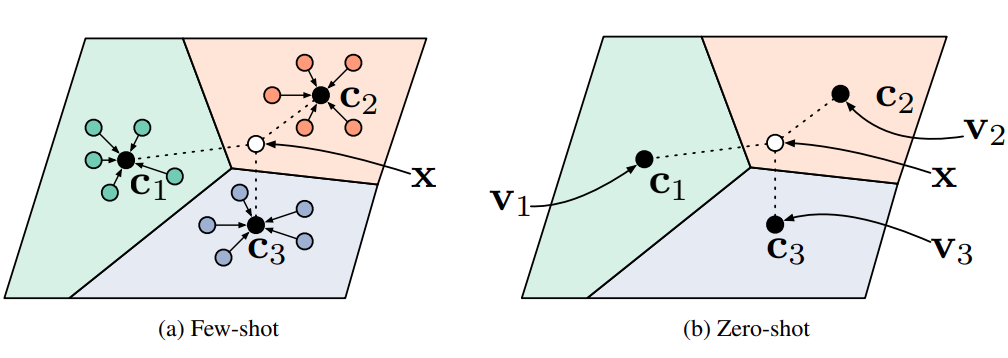}
    \caption{Prototypical Networks in the few-shot and zero-shot scenarios (source: \cite{snell2017prototypical}).}
    \label{fig:proto_net}
\end{figure}

Given a query $x$, a distribution over classes is produced by a softmax over negative distances to the prototypes:
\begin{equation}
    \label{eq:proto_dist}
    p_{\theta}(y = k \mid x) = \frac{\exp(-d(f_{\theta}(x), \textbf{c}_k))}{\sum_{k'} \exp( -d(f_{\theta}(x), \textbf{c}_{k'}))}
\end{equation}
where $d : \mathbb{R}^d \times \mathbb{R}^d \to [0, +\infty)$ is a differentiable Bregman divergence; here, the squared Euclidean distance
\begin{equation}
    \label{eq:euclidian_dist}
    d(f_{\theta}(x), \textbf{c}_k) = \| f_{\theta}(x) - \textbf{c}_k \|^2 .
\end{equation}
In the original formulation, training minimizes the negative log-probability $\mathcal{J}(\theta) = -\log p_{\theta}(y = k \mid x)$ of the true class; in this work we use Prototypical Networks only as an inference-time classification rule on top of the frozen, triplet-trained encoder.

\section{Training and Experimental Setup}
\label{section:training}

\subsection{Data}
\label{subsec:data}
\begin{figure}
\centering
\includegraphics[width=14.0 cm]{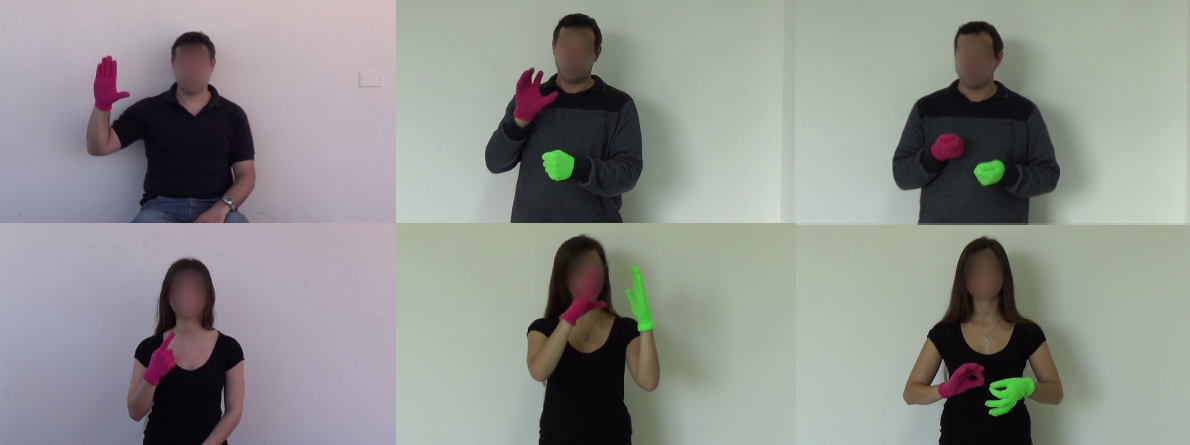}
\caption{Sample snapshots of the LSA64 dataset.}
\label{fig:lsa}
\end{figure}

We evaluate our method on the LSA64 dataset \cite{Ronchetti2016}, an Argentinian Sign Language dataset with 3200 videos in which 10 non-expert subjects perform 5 repetitions of 64 different sign classes, i.e., 50 videos per sign. The dataset was recorded in two sessions: the first outdoors under natural lighting, comprising 23 one-handed signs; the second indoors under artificial lighting, adding 22 two-handed and 19 one-handed signs. Subjects stood or sat in front of a white background wearing black clothes, and wore fluorescent-colored gloves to simplify hand segmentation and remove skin-color issues (Figure~\ref{fig:lsa}).

We note that the gloves may also ease hand key-point detection by MediaPipe; generalization to bare-handed signing is left as future work and is discussed in Section~\ref{subsec:limitations}.

The 64 classes are split into 48 classes (1--48) used to train the encoder and 16 held-out classes (49--64) used exclusively for few-shot evaluation; the test classes are therefore never seen during representation learning.

\subsection{Triplet Loss}
Given an input $\textbf{x}$, the Transformer encoder maps $\textbf{x}$ to a feature vector in $\mathbb{R}^d$. These representations must encode the class semantics: samples of the same sign should form clusters even when the videos differ (e.g., different signers), while samples of different signs should be distant even when recorded by the same person. To induce this property we train with the triplet loss.

\begin{figure}
\centering
\includegraphics[width=12.0cm]{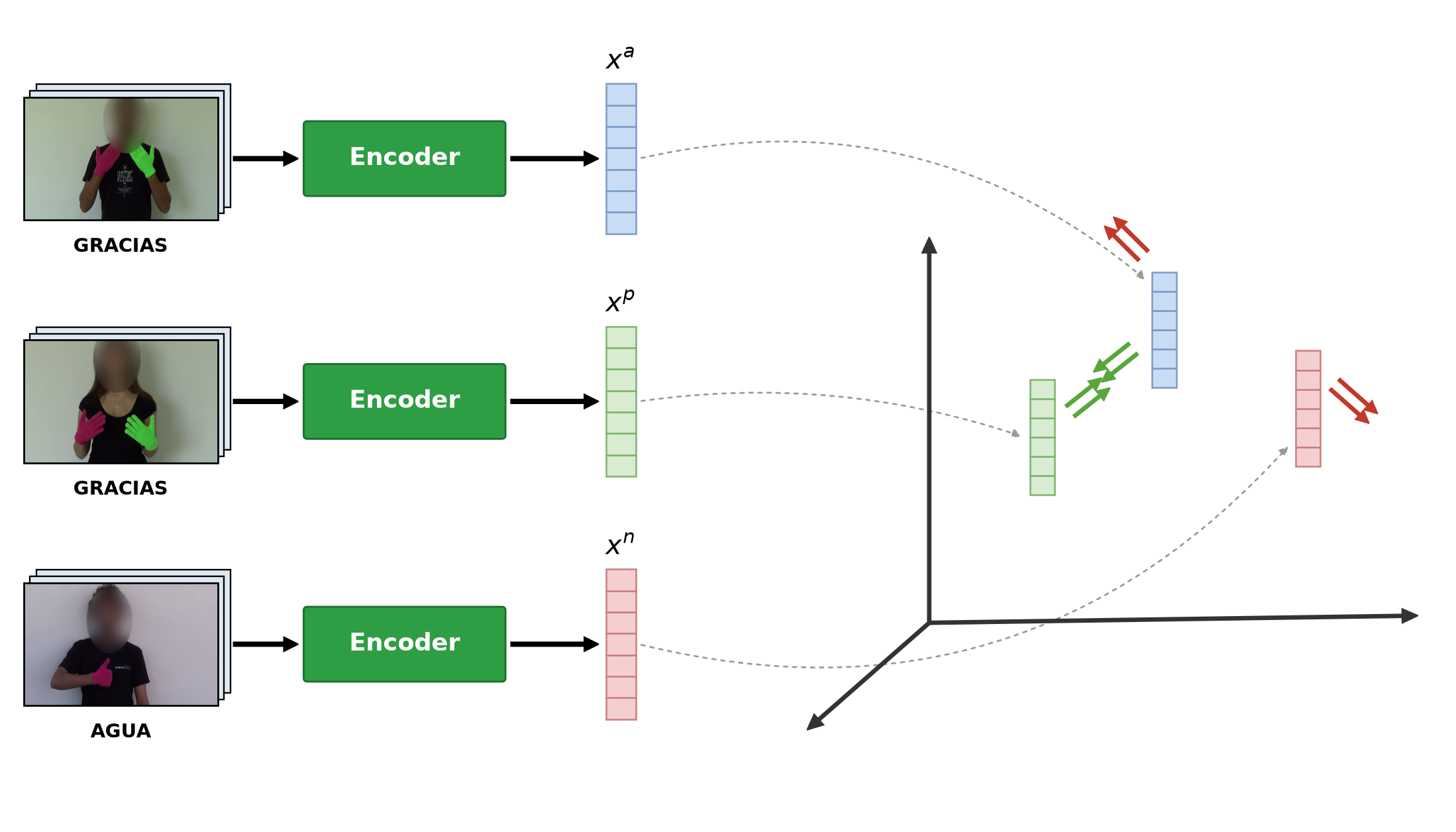}
\caption{Example of a triplet formed by an anchor, a positive, and a negative point. Each point is an encoding produced by the Transformer model. Training increases the distance between the anchor and the negative while decreasing the distance between the anchor and the positive.}
\label{fig:triplet}
\end{figure}

A triplet consists of three inputs: an \emph{anchor}, a \emph{positive} from the same class as the anchor, and a \emph{negative} from a different class. The loss maximizes the distance between anchor and negative while minimizing the distance between anchor and positive (Figure~\ref{fig:triplet}).

With embeddings $f(x^a)$, $f(x^p)$, $f(x^n)$ of the anchor, positive, and negative, and a margin $\alpha$ that prevents the collapse of all points to a single position, the loss is
\begin{equation}
    \label{eq:L}
    \mathcal{L}(x^a,x^p,x^n)=\max\!\left(\Vert f(x^a)-f(x^p)\Vert^2-\Vert f(x^a)-f(x^n)\Vert^2+\alpha,\; 0\right).
\end{equation}

Triplets are constructed by randomly choosing two distinct classes from the 48 training classes; two distinct samples of the first class serve as anchor and positive, and one sample of the second class as the negative. We sample 5{,}000 such triplets, re-drawing the set at every epoch. We use random (uniform) triplet sampling; online hard or semi-hard mining \cite{schroff2015facenet} is left for future work.

\subsection{Implementation Details}
\label{subsec:implementation}
Table~\ref{tab:hyper} lists all hyperparameters. The Transformer encoder input dimension is 225 (the flattened MediaPipe skeleton, Section~\ref{subsec:skeleton}); the model dimension is $d_{model}=128$ with $h=4$ attention heads, $n=2$ encoder layers, and a feed-forward dimension of 1024. Sequences are interpolated to a fixed length of $N=64$ frames. The final embedding dimension is 128.

\begin{table}[h!]
\centering
\begin{tabular}{ll}
\toprule
\textbf{Hyperparameter} & \textbf{Value} \\
\midrule
Input dimension ($3M$)            & 225 \\
Sequence length after interpolation ($N$) & 64 \\
Model dimension ($d_{model}$)     & 128 \\
Attention heads ($h$)             & 4 \\
Encoder layers ($n$)              & 2 \\
Feed-forward dimension            & 1024 \\
Dropout                           & 0.1 \\
Embedding dimension               & 128 \\
Positional-encoding base          & 10000 \\
Triplets per epoch                & 5000 \\
Triplet margin ($\alpha$)         & 1.0 \\
Optimizer                         & Adam \cite{kingma2014adam} \\
Learning rate                     & $1\times10^{-4}$ \\
Batch size (triplets)             & 64 \\
Epochs                            & 100 \\
\bottomrule
\end{tabular}
\caption{Hyperparameters of the model and training procedure (from \texttt{configs/default.yaml}).}
\label{tab:hyper}
\end{table}

The architecture was implemented in Python on 64-bit Ubuntu. The Transformer encoder was implemented with PyTorch \cite{paszke2019pytorch}; the $k$NN and cosine-similarity classifiers with scikit-learn \cite{pedregosa2011scikit}; and the Prototypical-Network evaluation directly in NumPy/PyTorch. All experiments in this paper were run on a single GPU with mixed-precision (AMP) training; the training code additionally supports multi-GPU and multi-node execution via PyTorch DistributedDataParallel. The released code includes the dataset download, key-point extraction, training, and evaluation scripts required to reproduce every number in this paper.

\subsection{Evaluation Protocols}
\label{subsec:protocols}
All evaluations use only the 16 held-out classes, unseen during encoder training. Unless stated otherwise, each protocol is repeated and we report the mean accuracy together with its dispersion.

\paragraph{Prototypical Networks ($K$-shot $N$-way).} An episode samples $N$ classes from the 16 test classes and $K$ support plus $Q$ query examples per class; prototypes are the support means and queries are assigned to the nearest prototype under squared Euclidean distance (Equations~\ref{eq:proto}--\ref{eq:euclidian_dist}). We use $Q=5$ query examples per class, run 600 episodes per configuration, and report mean accuracy with a 95\% confidence interval.

\paragraph{$k$-Nearest Neighbors.} For each repetition, $k$ support examples per test class are sampled as the reference set, and all remaining samples of the test classes are classified by majority vote among the $k$ nearest references (Euclidean distance). We set the number of support examples per class equal to $k$ so that, in a data-scarce scenario, all available reference points participate in the vote. Each configuration is repeated 40 times.

\paragraph{Cosine Similarity.} For each repetition, $n$ support examples per test class are sampled; a query is assigned the class of the single most similar support embedding (Equation~\ref{eq:cos}). Each configuration is repeated 40 times.

For the $k$NN and cosine protocols we additionally vary the number of classes used to train the encoder (from 10 to 45, in steps of 5) while keeping the number of triplets fixed, to quantify the effect of class diversity on the quality of the learned embedding space.

\section{Results}
\label{section:results}
The output of the training phase in Section~\ref{section:training} is an embedding space that can represent, in principle, any sign sharing structure with the training content. Few-shot methods can then classify unseen signs by comparing embeddings.

\subsection{Prototypical Networks}
We first measure few-shot classification performance with Prototypical Networks using the full model (encoder trained on all 48 training classes) under the $K$-shot $N$-way protocol of Section~\ref{subsec:protocols}. Table~\ref{tab:shots_1} reports the mean accuracy on the 16 unseen classes for every combination of $N\in\{4,6,8,10\}$ ways and $K\in\{1,5,10\}$ shots.

\begin{table}[h!]\centering
\small
\begin{tabular}{cccc}\toprule
\multirow{2}{*}{\textbf{Ways ($N$)}} &\multicolumn{3}{c}{\textbf{ACC (\%)}} \\\cmidrule{2-4}
&1-shot &5-shot &10-shot \\\midrule
4  &86.7\,$\pm$\,0.9 &94.3\,$\pm$\,0.6 &95.2\,$\pm$\,0.5 \\
6  &82.6\,$\pm$\,0.8 &91.8\,$\pm$\,0.5 &93.1\,$\pm$\,0.5 \\
8  &78.0\,$\pm$\,0.7 &89.2\,$\pm$\,0.5 &91.1\,$\pm$\,0.5 \\
10 &74.4\,$\pm$\,0.7 &87.8\,$\pm$\,0.5 &89.6\,$\pm$\,0.4 \\
\bottomrule
\end{tabular}
\caption{Prototypical-Network classification accuracy (\%, mean $\pm$ 95\% CI over 600 episodes) on the 16 unseen classes, for the full model (48 training classes), varying the number of ways $N$ and shots $K$.}\label{tab:shots_1}
\end{table}

Two consistent trends emerge. Accuracy grows with the number of support shots --- from 74.4\% (10-way 1-shot) to 89.6\% (10-way 10-shot) --- since larger support sets yield more reliable class prototypes. Conversely, accuracy decreases as the number of ways grows, because a larger candidate set makes each episode harder: the same model drops from 86.7\% at 4-way 1-shot to 74.4\% at 10-way 1-shot. Even in the hardest 10-way 1-shot setting, a single reference example per class suffices to recognize signs never seen during representation learning with 74.4\% accuracy, well above the 10\% chance level.

\subsection{$k$-Nearest Neighbors}
Table~\ref{table:knn} reports the mean accuracy ($\pm$ standard deviation over 40 repetitions) of the $k$NN protocol for $k=1$ to $8$, with the number of encoder training classes varying from 10 to 45.

\begin{table}[h!]
\centering
\scriptsize
\begin{tabular}{ccccccccc}
\toprule
\# train signs & $k=1$ & $k=2$ & $k=3$ & $k=4$ & $k=5$ & $k=6$ & $k=7$ & $k=8$ \\
\midrule
10 & 40.4\,$\pm$\,3.7 & 40.0\,$\pm$\,3.6 & 46.2\,$\pm$\,3.6 & 50.4\,$\pm$\,3.1 & 53.2\,$\pm$\,2.4 & 54.9\,$\pm$\,2.2 & 56.0\,$\pm$\,2.1 & 57.2\,$\pm$\,2.0 \\
15 & 48.6\,$\pm$\,3.1 & 46.7\,$\pm$\,4.0 & 54.6\,$\pm$\,2.5 & 58.3\,$\pm$\,2.5 & 61.0\,$\pm$\,2.4 & 62.7\,$\pm$\,2.5 & 63.8\,$\pm$\,2.1 & 65.3\,$\pm$\,1.9 \\
20 & 53.0\,$\pm$\,4.2 & 53.3\,$\pm$\,3.9 & 60.3\,$\pm$\,3.3 & 64.3\,$\pm$\,2.8 & 66.3\,$\pm$\,2.3 & 67.5\,$\pm$\,1.9 & 69.1\,$\pm$\,1.8 & 69.6\,$\pm$\,2.0 \\
25 & 54.1\,$\pm$\,3.9 & 51.7\,$\pm$\,3.1 & 60.2\,$\pm$\,3.2 & 63.9\,$\pm$\,3.0 & 66.4\,$\pm$\,2.5 & 68.1\,$\pm$\,2.6 & 69.3\,$\pm$\,2.3 & 70.2\,$\pm$\,2.3 \\
30 & 63.5\,$\pm$\,4.5 & 60.9\,$\pm$\,3.5 & 69.5\,$\pm$\,2.9 & 71.2\,$\pm$\,2.5 & 74.5\,$\pm$\,2.5 & 75.5\,$\pm$\,2.5 & 76.7\,$\pm$\,2.0 & 77.1\,$\pm$\,1.6 \\
35 & 59.9\,$\pm$\,4.6 & 57.5\,$\pm$\,3.1 & 66.4\,$\pm$\,3.2 & 70.1\,$\pm$\,2.8 & 72.4\,$\pm$\,3.0 & 74.3\,$\pm$\,2.2 & 76.1\,$\pm$\,2.4 & 76.8\,$\pm$\,2.1 \\
40 & 62.4\,$\pm$\,4.4 & 59.8\,$\pm$\,3.5 & 69.2\,$\pm$\,3.1 & 71.4\,$\pm$\,2.8 & 74.5\,$\pm$\,2.6 & 75.5\,$\pm$\,2.0 & 76.9\,$\pm$\,1.9 & 77.3\,$\pm$\,1.9 \\
45 & 64.5\,$\pm$\,4.6 & 62.5\,$\pm$\,3.7 & 71.9\,$\pm$\,3.3 & 74.0\,$\pm$\,2.6 & 76.8\,$\pm$\,2.3 & 77.5\,$\pm$\,2.1 & 78.9\,$\pm$\,1.8 & 79.5\,$\pm$\,1.8 \\
\bottomrule
\end{tabular}
\caption{$k$NN classification accuracy (\%, mean $\pm$ std over 40 repetitions) on the 16 unseen classes, varying the number of signs used to train the encoder; the number of support samples per class equals $k$.}
\label{table:knn}
\end{table}

\begin{figure}
\centering
\includegraphics[width=12.0cm]{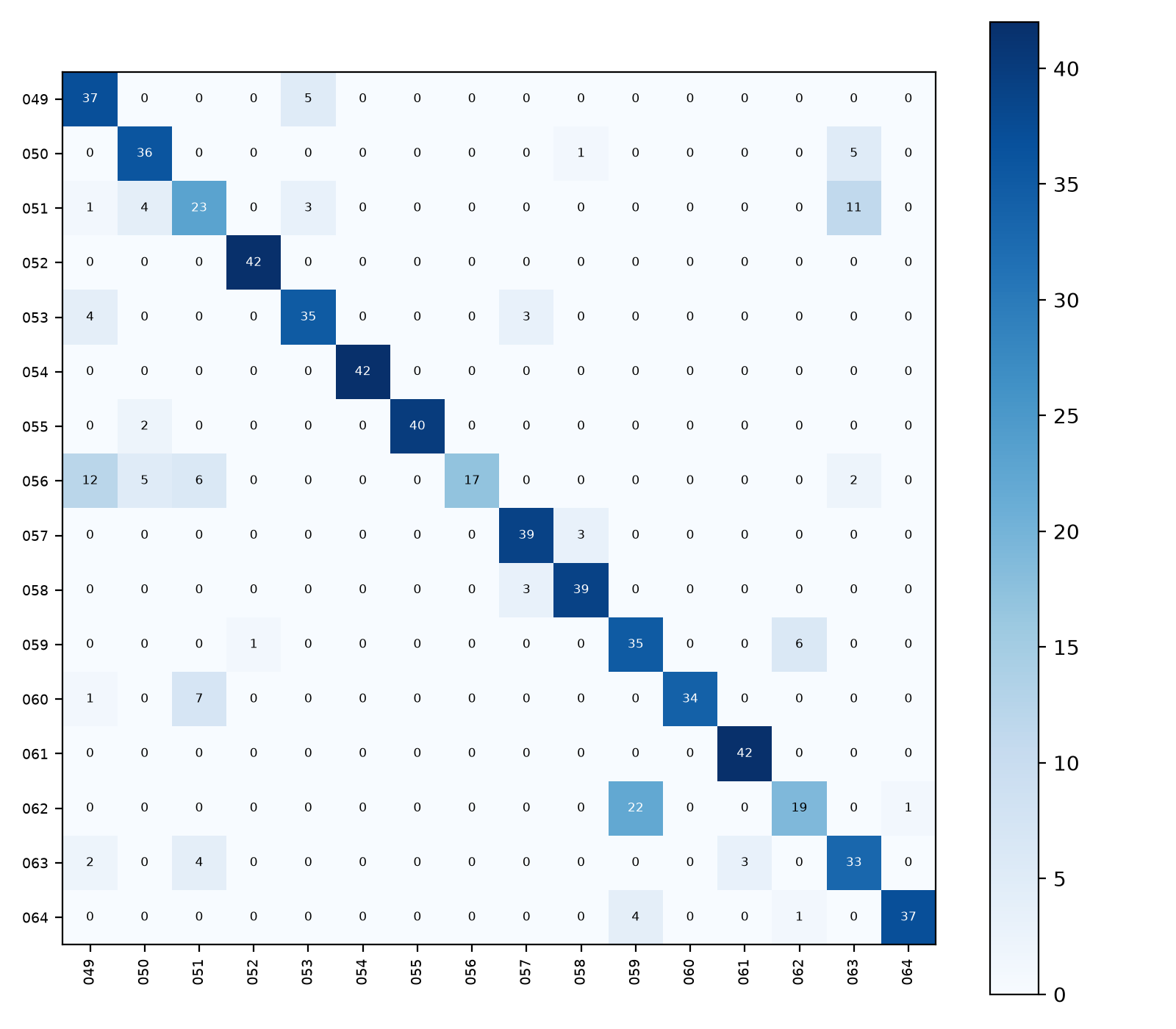}
\caption{Confusion matrix for $k$NN classification with $k=8$ and the full model (encoder trained on all 48 training signs), evaluated on the 16 unseen classes (049--064).}
\label{fig:knn_cm}
\end{figure}

Two factors improve accuracy. First, increasing the number of encoder training classes yields consistently better embeddings: at $k=8$, accuracy rises from 57.2\% with 10 training signs to 79.5\% with 45, confirming that class diversity during representation learning benefits classes never seen during training. Second, accuracy grows with $k$, as a larger support set (recall that the number of reference examples per class equals $k$) provides more evidence per vote. The exception is $k=2$, which frequently underperforms $k=1$ (e.g., 62.5\% vs.\ 64.5\% with 45 signs) because ties in a two-neighbor vote are resolved arbitrarily. Fluctuations across adjacent training-class settings (for instance, 35 signs falling slightly below 30) reflect the variance expected from the small 16-class test set. Training the encoder on all 48 classes (the full model) further raises the best $k=8$ accuracy to 81.6\%; its confusion matrix is shown in Figure~\ref{fig:knn_cm}.

\begin{figure}
\centering
\includegraphics[width=14.0cm]{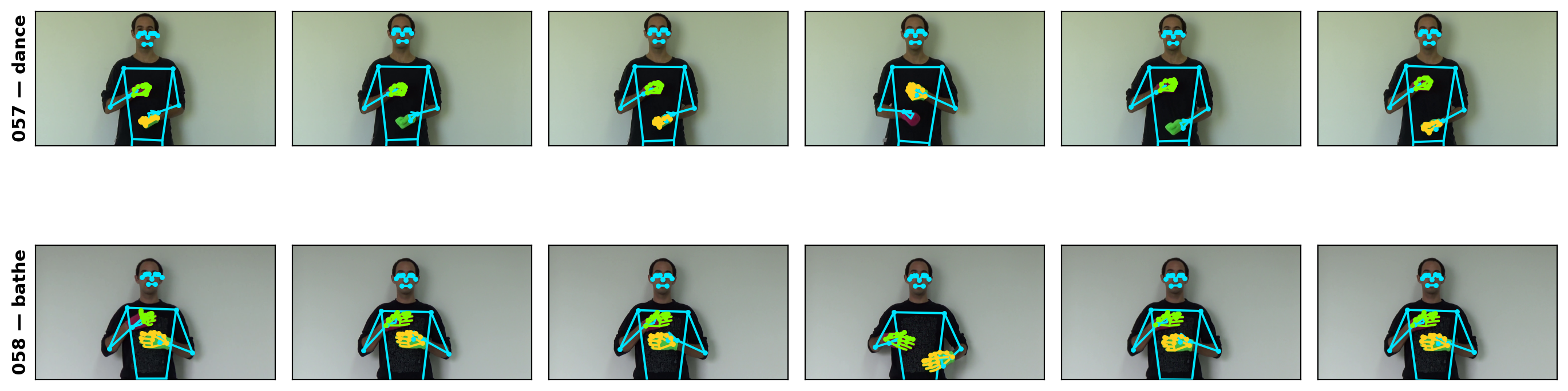}
\caption{Comparison between the signs for ``dance'' and ``bathe''. In some samples the model mislabels these signs because of their gestural similarity.}
\label{fig:compare}
\end{figure}

The main diagonal concentrates the correct predictions. Off-diagonal entries occur mostly for signs that share gestural semantics; the encoder maps them close together in the embedding space and the classifier mislabels them. For example, sign 057 (\emph{dance}) and sign 058 (\emph{bathe}) have similar gestures, as shown in Figure~\ref{fig:compare}.

\subsection{Cosine Similarity}
The cosine-similarity experiments use the same configuration as the $k$NN experiment: encoder training classes vary from 10 to 45 in steps of 5, and the number of support samples per unseen class, $n$, varies from 1 to 8. The prediction is the class of the single most similar support embedding. Each combination is evaluated over 40 repetitions (Table~\ref{table:cos}).

\begin{table}[h!]
\centering
\scriptsize
\begin{tabular}{ccccccccc}
\toprule
\# train signs & $n=1$ & $n=2$ & $n=3$ & $n=4$ & $n=5$ & $n=6$ & $n=7$ & $n=8$ \\
\midrule
10 & 40.3\,$\pm$\,3.7 & 50.4\,$\pm$\,2.8 & 55.7\,$\pm$\,3.2 & 60.2\,$\pm$\,2.9 & 64.0\,$\pm$\,2.8 & 66.6\,$\pm$\,2.4 & 68.8\,$\pm$\,2.4 & 70.5\,$\pm$\,2.2 \\
15 & 48.4\,$\pm$\,3.1 & 58.6\,$\pm$\,2.8 & 64.4\,$\pm$\,3.3 & 68.5\,$\pm$\,3.3 & 71.8\,$\pm$\,3.1 & 74.8\,$\pm$\,2.7 & 77.1\,$\pm$\,2.1 & 79.0\,$\pm$\,1.7 \\
20 & 52.4\,$\pm$\,4.3 & 62.5\,$\pm$\,3.3 & 67.3\,$\pm$\,2.4 & 71.5\,$\pm$\,2.7 & 74.8\,$\pm$\,2.4 & 76.9\,$\pm$\,2.3 & 78.7\,$\pm$\,2.2 & 80.1\,$\pm$\,2.0 \\
25 & 54.5\,$\pm$\,3.7 & 64.1\,$\pm$\,3.5 & 69.8\,$\pm$\,2.8 & 73.9\,$\pm$\,2.6 & 76.5\,$\pm$\,2.5 & 78.6\,$\pm$\,2.0 & 80.2\,$\pm$\,1.7 & 81.7\,$\pm$\,1.8 \\
30 & 63.5\,$\pm$\,4.4 & 72.0\,$\pm$\,3.0 & 76.3\,$\pm$\,2.8 & 79.6\,$\pm$\,2.4 & 82.5\,$\pm$\,2.1 & 84.6\,$\pm$\,1.9 & 86.1\,$\pm$\,1.7 & 87.4\,$\pm$\,1.8 \\
35 & 59.6\,$\pm$\,4.5 & 68.3\,$\pm$\,3.6 & 73.7\,$\pm$\,2.5 & 77.6\,$\pm$\,2.6 & 80.4\,$\pm$\,2.3 & 82.3\,$\pm$\,2.3 & 83.8\,$\pm$\,1.9 & 85.0\,$\pm$\,1.9 \\
40 & 62.2\,$\pm$\,4.3 & 71.3\,$\pm$\,3.1 & 76.7\,$\pm$\,2.6 & 79.9\,$\pm$\,2.2 & 82.6\,$\pm$\,2.4 & 84.5\,$\pm$\,2.2 & 86.0\,$\pm$\,2.3 & 87.1\,$\pm$\,2.2 \\
45 & 64.3\,$\pm$\,4.5 & 73.6\,$\pm$\,3.4 & 78.1\,$\pm$\,2.3 & 80.9\,$\pm$\,2.2 & 83.5\,$\pm$\,2.2 & 85.4\,$\pm$\,1.9 & 86.6\,$\pm$\,1.8 & 87.9\,$\pm$\,1.7 \\
\bottomrule
\end{tabular}
\caption{Cosine-similarity classification accuracy (\%, mean $\pm$ std over 40 repetitions) on the 16 unseen classes, varying the number of signs used to train the encoder and the number of support samples per class, $n$.}
\label{table:cos}
\end{table}

\begin{figure}
\centering
\includegraphics[width=12.0cm]{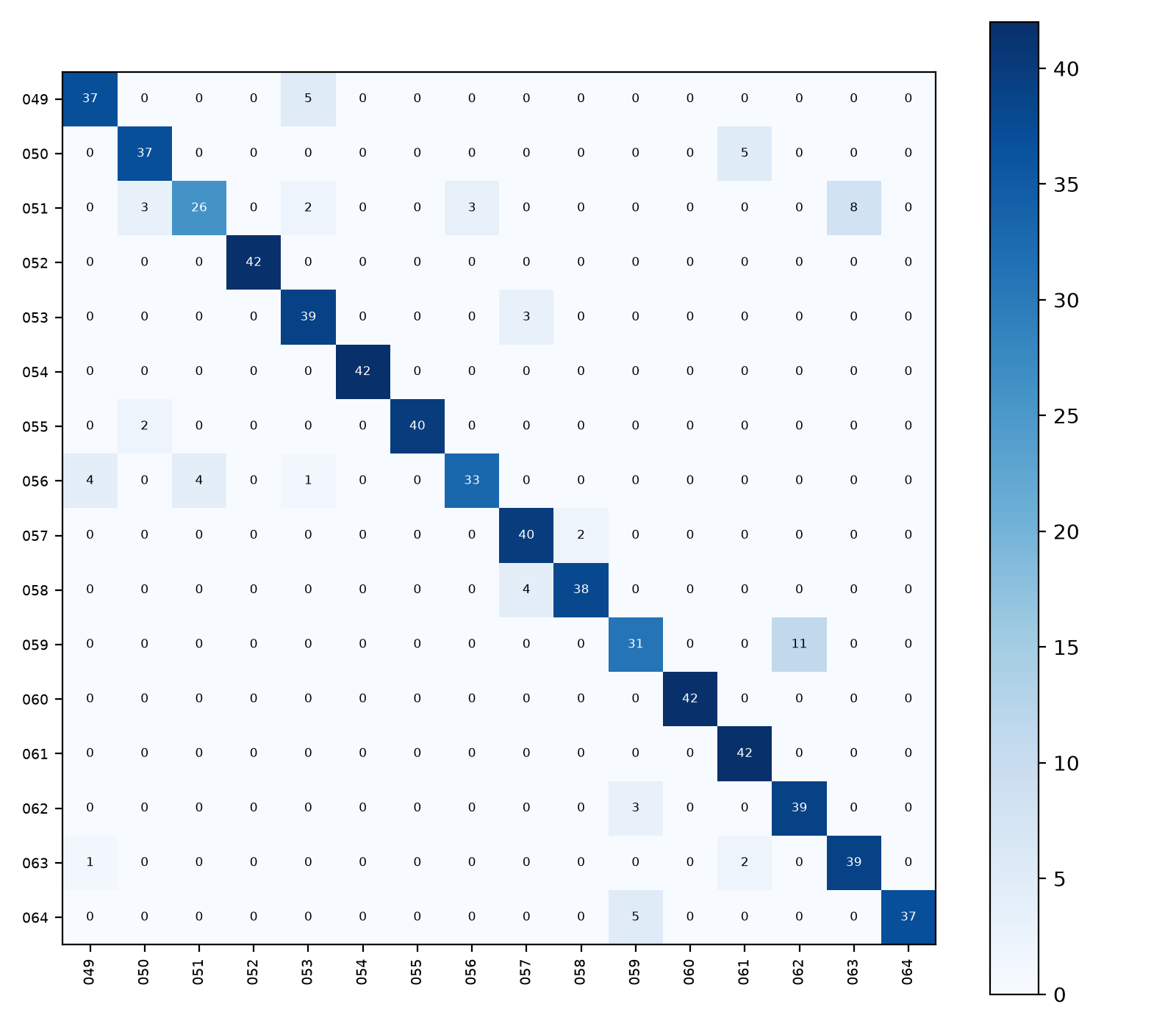}
\caption{Confusion matrix for cosine-similarity classification with $n=8$ and the full model (encoder trained on all 48 training signs), evaluated on the 16 unseen classes (049--064).}
\label{fig:cos_cm}
\end{figure}

Cosine similarity follows the same two trends as $k$NN --- accuracy increases with both the number of training signs and the number of support examples $n$ --- but is uniformly stronger. At 45 training signs it reaches 87.9\% with $n=8$, roughly eight points above the corresponding $k$NN configuration (79.5\%), and it already surpasses 80\% with only three support examples. The advantage is largest in the low-shot regime: with a single reference example ($n=1$) the two rules are comparable, but cosine similarity separates from $k$NN as $n$ grows, indicating that the triplet-trained embedding space is better organized by angular (direction) similarity than by Euclidean majority voting. The full model (all 48 training classes) attains the best overall result, 88.4\%\ with $n=8$; its confusion matrix is shown in Figure~\ref{fig:cos_cm}.

\subsection{Comparison with Prior Work}
\label{subsec:comparison}
Table~\ref{tab:sota} contrasts our results with published results on LSA64. A direct comparison must be made with care: prior work performs \emph{supervised, closed-set} classification over all 64 classes, using many labeled examples per class and evaluating on classes seen during training, whereas our protocol evaluates on 16 classes that are \emph{never seen} during representation learning, using at most 8 labeled examples per class at inference time. Supervised signer-dependent results, such as the 95.95\% of the original dataset paper \cite{Ronchetti2016} or the 98.09\% obtained by a four-stream skeletal LSTM \cite{konstantinidis2018sign}, should therefore be read as an upper bound under much stronger assumptions. A more instructive reference point is the signer-\emph{independent} baseline of Marais et al.\ \cite{marais2022investigating}, whose accuracy drops to 74.22\% (from 97.03\% in the signer-dependent setting): generalizing across signers is difficult on LSA64 even with full supervision and all classes seen. Our protocol targets an orthogonal and complementary form of generalization --- holding out entire sign classes rather than signers --- and reaches 88.4\% on 16 unseen classes with only eight references per class, without any retraining. Combining the two axes (unseen classes \emph{and} unseen signers) remains future work (Section~\ref{subsec:limitations}).

\begin{table}[h!]
\centering
\small
\begin{tabular}{lllc}
\toprule
\textbf{Method} & \textbf{Setting} & \textbf{Classes at test} & \textbf{ACC (\%)} \\
\midrule
Ronchetti et al.\ \cite{Ronchetti2016} (HMM/GMM sub-classifiers) & supervised, signer-dep. & 64 seen & 95.95 \\
Konstantinidis et al.\ \cite{konstantinidis2018sign} (skeletal LSTM, late fusion) & supervised, signer-dep. & 64 seen & 98.09 \\
Marais et al.\ \cite{marais2022investigating} (InceptionV3-GRU) & supervised, signer-indep. & 64 seen & 74.22 \\
\midrule
Ours (ProtoNet, 8-way 10-shot) & few-shot & 16 unseen & 91.1\,$\pm$\,0.5 \\
Ours (cosine, $n=8$, full model) & few-shot & 16 unseen & 88.4\,$\pm$\,1.8 \\
\bottomrule
\end{tabular}
\caption{Comparison with prior work on LSA64. Prior methods are fully supervised and evaluate on all 64 classes seen during training; our method evaluates on 16 classes never seen during representation learning, with at most 8 labeled examples per class. Signer-dependent baselines share signers between train and test; Marais et al.\ report a signer-independent split.}
\label{tab:sota}
\end{table}

\subsection{Visualizing the Embedding Vector Space}
It is infeasible to visualize the high-dimensional feature space produced by the Transformer directly. To gain intuition about the distribution of the embedding space, we use t-SNE (t-Distributed Stochastic Neighbor Embedding) \cite{van2008visualizing}, a manifold approach to dimensionality reduction. t-SNE converts affinities between data points into probabilities: affinities in the original space are modeled by Gaussian joint probabilities and affinities in the low-dimensional map by Student's t-distributions, which makes the technique particularly sensitive to local structure.

\begin{figure}
\centering
\includegraphics[width=14.0cm]{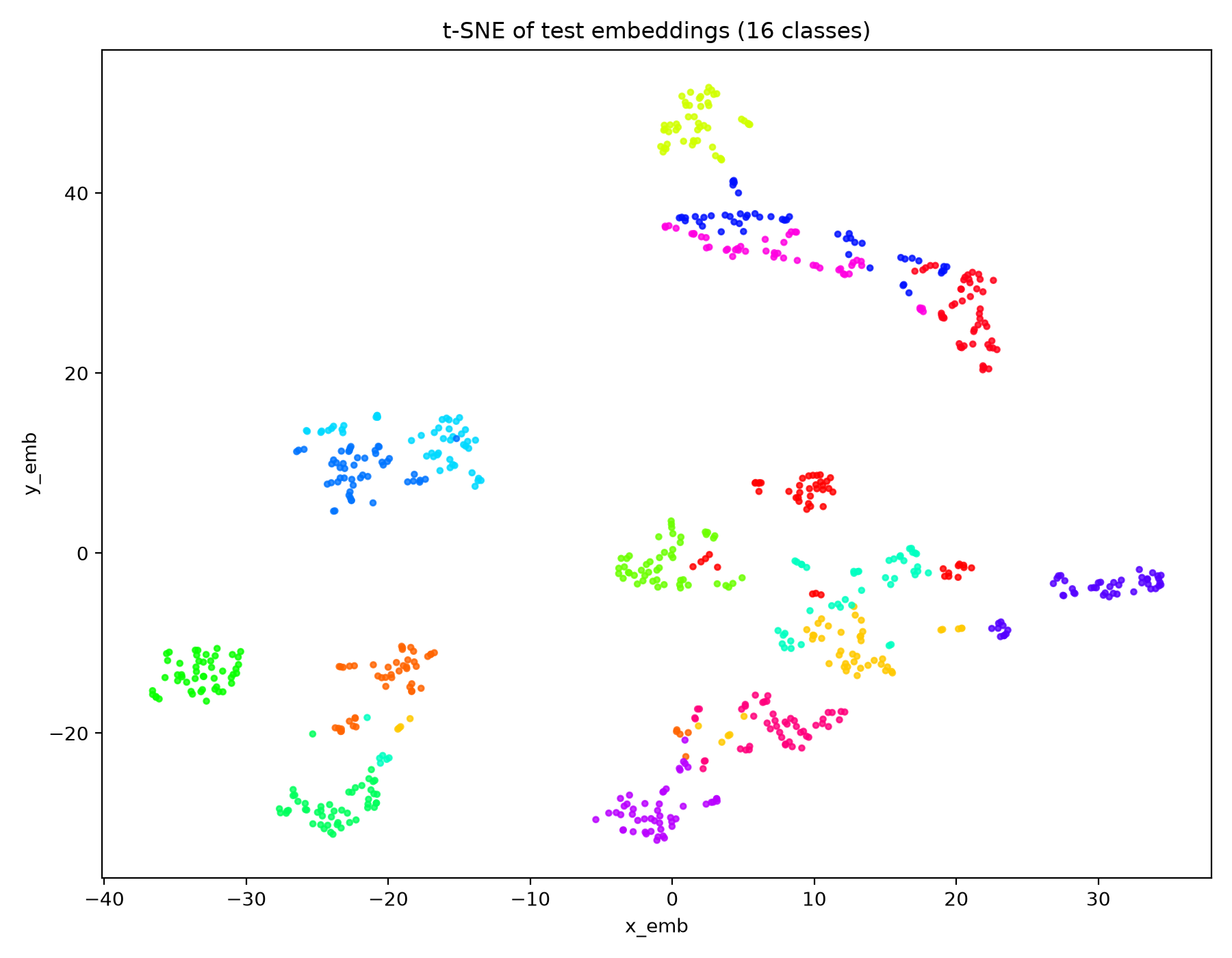}
\caption{t-SNE visualization of test data (16 unseen classes) in the embedding space. Each color represents a class and each point a sample.}
\label{fig:tsne_test}
\end{figure}

Figure~\ref{fig:tsne_test} shows the two-dimensional t-SNE projection of the embeddings of the 16 unseen test classes (50 samples per class); each color represents a class and each point a sample. Crucially, none of these classes was seen during representation learning.

The projection reveals well-separated, compact clusters for most unseen classes, indicating that the encoder maps semantically equivalent signs close together even though they were never used to train it --- the property that makes few-shot comparison of embeddings effective. A few clusters lie close to or partially overlap one another, which is consistent with the off-diagonal confusions observed in the confusion matrices (Figures~\ref{fig:knn_cm} and \ref{fig:cos_cm}) and corresponds to signs that share gestural structure.

\subsection{Limitations}
\label{subsec:limitations}
Three limitations should be noted. First, LSA64 subjects wear fluorescent gloves, which may ease hand key-point detection; generalization to bare-handed signing has not been evaluated. Second, our class split holds out signs but not signers: all 10 subjects appear in both encoder training and evaluation, so signer-independent generalization is untested. Third, LSA64 comprises isolated signs recorded under controlled conditions; performance on continuous signing and in-the-wild footage remains an open question. The released code supports signer-held-out splits, which we plan to explore in future work.

\section{Conclusions}
\label{section:concl}
We have shown that a Transformer encoder trained with a contrastive (triplet) objective can generate useful representations of body key-point sequences. The model extracts features from the input sequence in parallel without losing temporal information, thanks to the positional encoding. This hypothesis was tested by performing few-shot classification of signs never seen during representation learning, using $k$NN, cosine similarity, and Prototypical Networks, achieving accuracies of up to 95.2\% (ProtoNet, 4-way 10-shot) and 88.4\,$\pm$\,1.8\% under the harder 16-way cosine-similarity protocol with only eight reference examples per class. The positive correlation between accuracy and the number of training classes --- clearly visible across the $k$NN and cosine sweeps, where accuracy rises monotonically from 10 to 45 training signs --- suggests that class diversity benefits the model even with a modest number of samples per class (50 in LSA64). The concentration of predictions on the main diagonal of the confusion matrices indicates consistent per-class behavior, with confusions restricted to gesturally similar signs, which lie close together in the embedding space. Finally, t-SNE visualizations of unseen classes suggest that the model clusters new similar signs while keeping different ones apart.

As future work, datasets with a larger variety of classes and different sign languages can be used; signer-independent evaluation and online triplet mining can be incorporated; and the interpretability of the Transformer can be analyzed by visualizing internal attention weights.

\section*{Reproducibility Statement}
All code, configuration files, dataset download and key-point extraction scripts, and evaluation protocols used in this paper are publicly available at \url{https://github.com/esdrascosta/slr-fewshot}. Training supports distributed execution via PyTorch DDP, and a synthetic-data smoke test allows end-to-end verification of the pipeline without downloading the dataset. All tables in Section~\ref{section:results} are produced directly by the evaluation scripts.

\section*{Acknowledgments}
This research was partially funded by Lenovo, as part of its R\&D investment under Brazil's Informatics Law. The authors want to acknowledge the support of Lenovo R\&D and CESAR D\&O.

%Bibliography
\bibliographystyle{unsrt}
\bibliography{references}

\end{document}